\documentclass[letterpaper, 10 pt, conference]{ieeeconf}  

\IEEEoverridecommandlockouts                              

\usepackage{caption}

\usepackage{amsmath}
\usepackage{amsfonts}
\usepackage{caption}
\usepackage{array}
\usepackage{float}
\usepackage{mathtools}
\usepackage{xcolor}
\usepackage{hyperref}

\title{\LARGE \bf
Tracking and Following a Suspended Moving Object using Camera-Based Vision System
}

\author{Michele Ambrosino$^{1}$, Manar Mahmalji$^{1}$,  Nicolás Bono Rosselló$^{1}$, and Emanuele Garone$^{1}$
\thanks{This research has been funded by The Brussels Institute for Research and Innovation (INNOVIRIS) of the Brussels Region through the Applied PHD grant: Brickiebots - Robotic Bricklayer: a multi-robot system for sand-lime blocks masonry (réf :  19-PHD-12)}
\thanks{$^{1}$Service d’Automatique et d’Analyse des Systèmes, Université Libre de Bruxelles, Brussels, Belgium.
        {\tt\small Manar.Mahmalji@ulb.be}
        {\tt\small Michele.Ambrosino@ulb.ac.be}
        {\tt\small nbonoros@ulb.ac.be}
        {\tt\small egarone@ulb.ac.be}}
}

\begin{document}

\maketitle
\thispagestyle{empty}
\pagestyle{empty}

\begin{abstract}
When robots are able to see and respond to their surroundings, a whole new world of possibilities opens up. To bring these possibilities to life, the robotics industry is increasingly adopting camera-based vision systems, especially when a robotic system needs to interact with a dynamic environment or moving target. However, this kind of vision system is known to have low data transmission rates, packet loss during communication and noisy measurements as major disadvantages. These problems can perturb the control performance and the quality of the robot-environment interaction. To improve the quality of visual information, in this paper, we propose to model the dynamics of the motion of a target object and use this model to implement an Extended Kalman Filter based on Intermittent Observations of the vision system. The effectiveness of the proposed approach was tested through experiments with a robotic arm, a camera device in an eye-to-hand configuration, and an oscillating suspended block as a target to follow.

\end{abstract}

\section{INTRODUCTION}

Vision plays an important role in a robotic system, as it can be used to obtain information about the environment where the robot operates \cite{computer_vision_systems}. This information can be exploited by the control system of the robot to ensure a correct interaction with the surrounding environment. Control based on visual measurements is known as Visual Servoing (VS).

The first important feature in VS applications is the camera-robot configuration \cite{flandin2000eye}: 1) eye-to-hand, the camera is placed in a fixed pose w.r.t the base frame of the robotic arm and faces the working field of the end-effector of the robot; and 2) eye-in-hand, the camera is mounted on the robotic arm, and moves along with it \cite{kudryavtsev2018eye}.

The second feature of VS applications is the target with which the robot must interact. Typically, in VS applications, the targets are static which facilitates the controller design. In \cite{saxena2008robotic}, the authors propose a learning algorithm to grab an unknown object. In \cite{ma2020robotic}, a VS application is developed with the goal of reaching and grasping a static object. A predictive controller is used in \cite{gangloff2002visual} to follow a 3-D static profile. However, a more challenging application is when the targets are moving \cite{agah2004line}. As a matter of fact, in practical applications the use of moving objects is becoming increasingly popular in both research \cite{marturi2019dynamic} and industries \cite{wang2020autonomous}. A dynamic VS is discussed in \cite{9812081}, where an eye-in-hand configuration is used to interact with a moving target. In \cite{6696335}, the authors address the problem of the disturbance on the visual feature dynamics due to the target motion. A pick and place task with a moving object is analysed in \cite{wong2022moving}. In all the aforementioned papers, the objects with which the robot must interact are relatively simple (e.g. plates, bars, cubes). Therefore, the dynamic model of the object is often not considered during the control design. It is also worth noting that, especially in the case of moving targets, problems such as low camera frame rates, packet loss and measurement noise can lead to incorrect object tracking (in some cases even unwanted impacts between robot and target).

\begin{figure}[h]
\centering
\includegraphics[width=0.5\columnwidth]{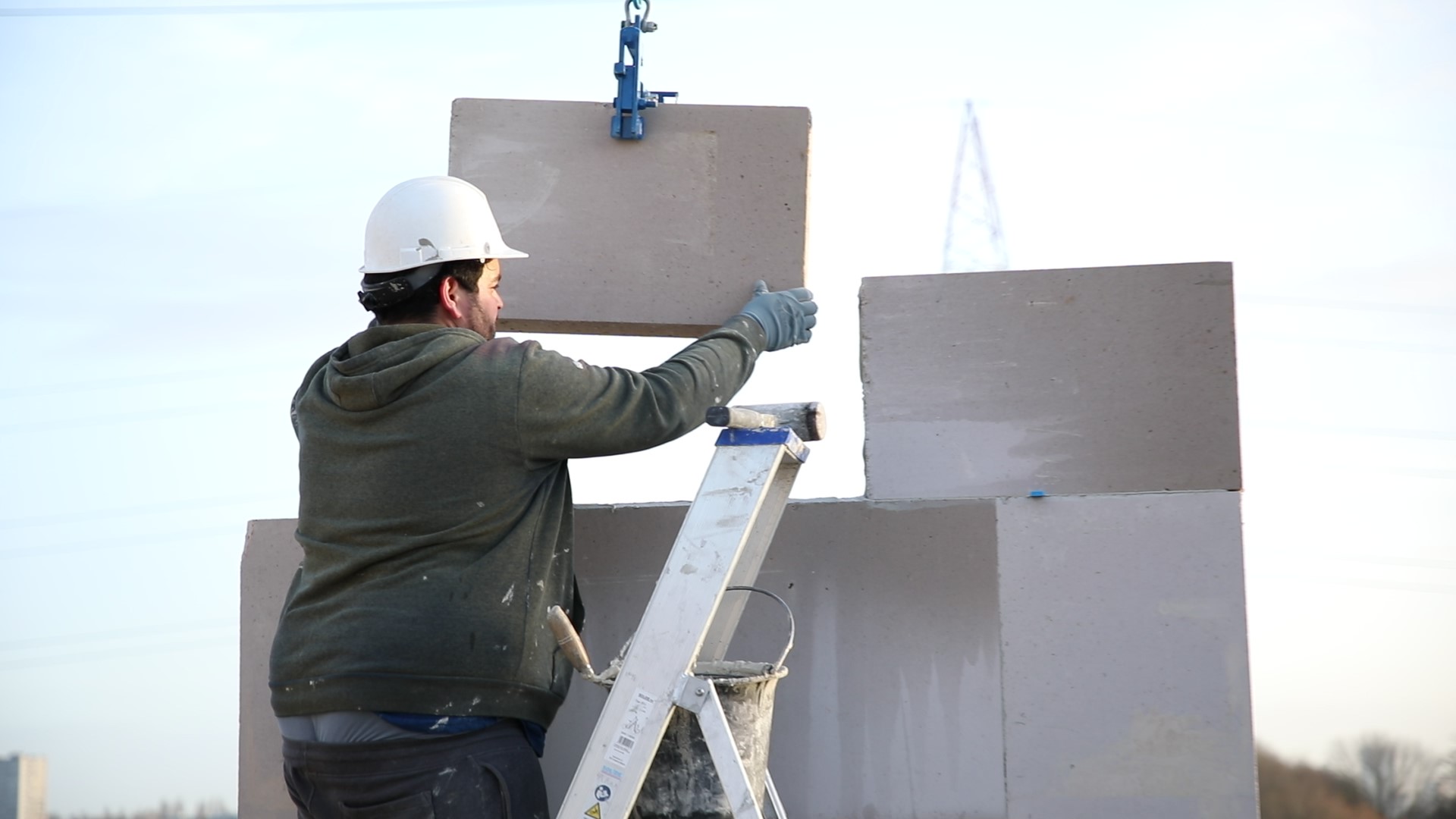}
\caption{Laying of a suspended block. }
\label{fig:mason} 
\end{figure}

In this work, we use the laying of a suspended block as a case study, shown in Fig.~\ref{fig:mason}. In this type of construction activity, the positioning of the block is divided into three sub-tasks: 1) the mason follows the oscillations of the block, 2) the mason grasps the block, and 3) the mason positions the block in its final position. In this paper, we focus on Task 1, while Tasks 2 and 3 represent future exploitaions of this work. To perform this same type of task with a robotic arm a vision system is needed to estimate the pose of the suspended object. The camera is placed in a fix pose w.r.t the base of the robot, therefore an eye-to-hand configuration is chosen to perform the desired task. However, as already mentioned, in the case of moving targets, in order to avoid unwanted impacts between the robot and the object, the typical problems of vision systems (such as low camera frame rate, loss of packets, etc.) must be addressed. 

Therefore, the aim of this work is to achieve accurate tracking of a moving object by a robotic arm. To this end, we use an Extended Kalman Filter (EKF) with Intermittent Observations. This estimator makes use of the dynamic model of the moving object and the measured pose of the object by a camera device to estimate the pose of the target at a higher rate than that of the camera and to feed the controller of the robot with a smooth and noise-free signal.        
The remainder of this work is organized as follows. Section~\ref{sec:probl} provides the problem formulation. In Section~\ref{pose_estimation}, the camera-to-robot calibration procedure is discussed. Section~\ref{sec:EKF} presents the proposed estimator and the problems related to the visual application. In Section~\ref{sec:results}, experimental results will be shown. Section~\ref{sec:conc} concludes the paper.

\section{PROBLEM STATEMENT}\label{sec:probl}

The system under investigation consists of a robotic arm that receives visual information from a camera device with the aim of following the movement of a suspended object. This object can be described as a 5-Dof pendulum, and its dynamic model will be explained in detail. 

\subsection{Description of the system}

\begin{figure}[h]
\centering
\includegraphics[width=0.8\columnwidth]{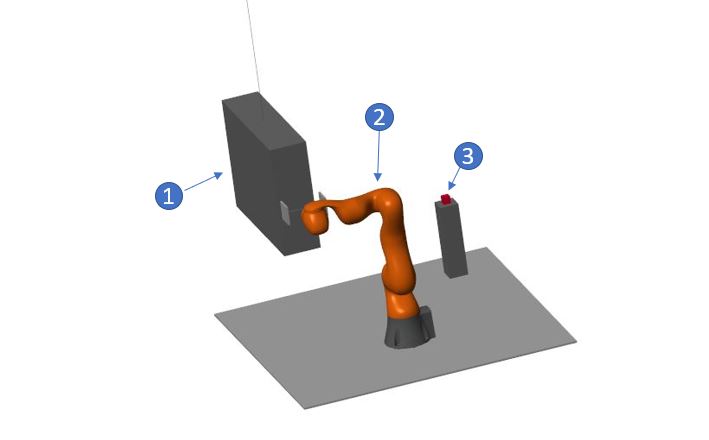}
\caption{Schematic view of the set-up. 1) Suspended block. 2) Robotic Arm. 3) Camera. }
\label{fig:scheme} 
\end{figure}

The overall system used in this work consists of three units: 1) a suspended object; 2) a robotic arm and 3) a camera device. A schematic view of the system is shown in Fig.~\ref{fig:scheme}. In this section, we briefly describe the dynamic model of the robotic arm and its controller. For an n-DoF manipulator, the dynamic model of a robotic arm can be written using the \textit{Lagrange formulation in the joint space} as \cite{sicilianos} 
\begin{equation}\label{eq:robot_model}
    \small
    B(q)\ddot{q} + C(q,\dot{q})\dot{q} + g(q) = \tau - J(q)h_e,
\end{equation}
where, $q$, $\dot{q}$, and $\ddot{q}$ $\in\mathbb{R}^{n}$ are the generalized joint positions, velocities and accelerations, respectively. The matrices $B(q) \in\mathbb{R}^{n\times n}$, $ C(q,\dot{q})\in\mathbb{R}^{n\times n}$, and $g(q) \in\mathbb{R}^{n}$ represent the inertia, centripetal-Coriolis, and gravity term, respectively. Moreover, $h_e \in\mathbb{R}^6$ 
represents the vector of the forces generated by contact with the environment, $J(q) \in\mathbb{R}^{6\times n}$ is the manipulator's geometric Jacobian, and $\tau \in\mathbb{R}^{n}$ is the vector of the control input.

\medskip

As mentioned earlier, the aim is to track, with the help of the camera, the oscillations of the suspended block to send commands to the robotic arm to follow the movement of the block. First, the camera reads the pose of the block (i.e. the orientation and the position of the center of gravity (COG) of the block). Then, this pose is translated into a convenient pose for tracking the oscillations of the block (i.e. the side of the suspended object), which represents the desired robot's end-effector pose $x_d$ to be reached. To reach this pose, the 
robot control input $\tau$ in \eqref{eq:robot_model} is designed as \cite{cntrl1} 

\begin{equation}\label{tau}
    \small
    \tau = \tau_{cmd} + \tau_{ns}
\end{equation}

where

\begin{equation}
    \small
    \tau_{cmd} = J^T(q)K_{P}\Tilde{x} - J^T(q)K_{D}J(q)\dot{q} + g(q),
\end{equation}
    with $\Tilde{x} = x_d - x_e$, $x_d$ being the  desired end-effector pose and $x_e$ being the current end-effector pose. $K_{p}$ and $K_{d}$ are  $(6\times6)$ symmetric positive definite matrices. And,
\begin{equation}\label{tau_ns}
   \small
   \tau_{ns} = (I_n J^T(q)\Bar{J}^T(q))w_0,
\end{equation}

where $\Bar{J}(q)$ is the generalized inverse of the Jacobian matrix, and $w_0$ is an arbitrary law used to achieve a secondary task. In this work, the secondary task aims at keeping the joint of the robots far from their mechanical limits to avoid that the robot will stop during the execution of the task. 

\medskip

It is worth noting that the control inputs \eqref{tau_ns} are needed since the robotic arm used in this work is a redundant manipulator. Moreover, to deal with orientations singularity, a non-minimal representation, such as Unit Quaternions, is used for orientation descriptions (see \cite{quat} for more detail on this subject).  

\subsection{Model of the suspended object}

\label{dyn_model}
\begin{figure}[h]
\centering
\includegraphics[width=0.25\columnwidth]{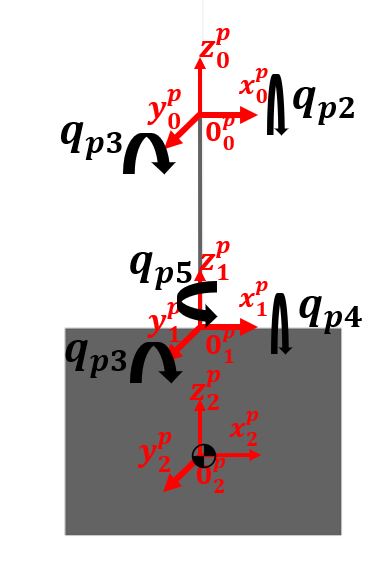}
\caption{Suspended block configuration.}
\label{fig:block} 
\end{figure}

The dynamics of the suspended block can be treated as that of a 5-DoF non-actuated pendulum, see Fig.~\ref{fig:block}. In particular, the configuration of the suspended object can be described by five variables, $ q_{p}\in\mathbb{R}^{5}$, with $q_p = \begin{bmatrix} q_{p1}, q_{p2},q_{p3},q_{p4},q_{p5}\end{bmatrix}^T$. Where $q_{p1}$ and $q_{p2}$ are the radial sway and the tangential pendulation respectively, and $q_{p3}$, $q_{p4}$, $q_{p5}$ are the orientations of the block \textit{w.r.t.} the cable. It is possible to show that the dynamic model of the suspended object can be written in a compact matrix form analogous to \eqref{eq:robot_model} as

\begin{equation}\label{eq:obj_model}
    \small
    B_p(q_{p})\ddot{q_{p}} + C_{p}(q_{p},\dot{q_{p}})\dot{q_{p}} + g_p(q_{p}) = 0,
\end{equation}
where $B_p(q_p) \in\mathbb{R}^{5\times 5}$, $ C_p(q_p,\dot{q_p})\in\mathbb{R}^{5\times 5}$, and $g_p(q_p) \in\mathbb{R}^{5}$. Defining the state vector as $x = [x_1^\textnormal{T}\ x_2^\textnormal{T}]^\textnormal{T}\in\mathbb{R}^{10}$, where $x_1 = q_{p}$ and $x_2 = \dot{q_{p}}$, we can rewrite equation \eqref{eq:obj_model} in the state-space representation as

\begin{equation}\label{eq:model}
\small
\dot{x} = f(x) =
\begin{bmatrix}
{x}_{2}\\
-B_P^{-1}({x}_{1})[C_p({x}_{1},{x}_{2})x_2+g_p({x}_{1})]
\end{bmatrix}.
\end{equation}

In this paper, we use a camera vision system to collect information about the states of the suspended object. In particular, the camera computes the pose of the block and gives as output the position and the orientation of the COG of the block w.r.t. a fixed frame (the frame $O^p_0$ in Fig~\ref{fig:block}). In this work, we treat the output of the camera as a non-linear output function $h(x)$ of the state-space model of \eqref{eq:model}. Therefore, the model of the suspended object in a state-space representation with the output equation can be written in the form \footnote{The mathematical expression of the state-space representation with the output equation is available in an open source GitHub repository, please visit \url{https://github.com/MikAmb95/EKF.git}.}:

\begin{equation}\label{eq:Tc}
    \small
    \begin{cases}
        \dot{x} &= f(x) \\
        y &= h(x).
    \end{cases}
\end{equation}

\section{POSE ESTIMATION WITH VISION SYSTEM} \label{pose_estimation}
\begin{figure}[H]
\centering
\includegraphics[width=0.55\columnwidth]{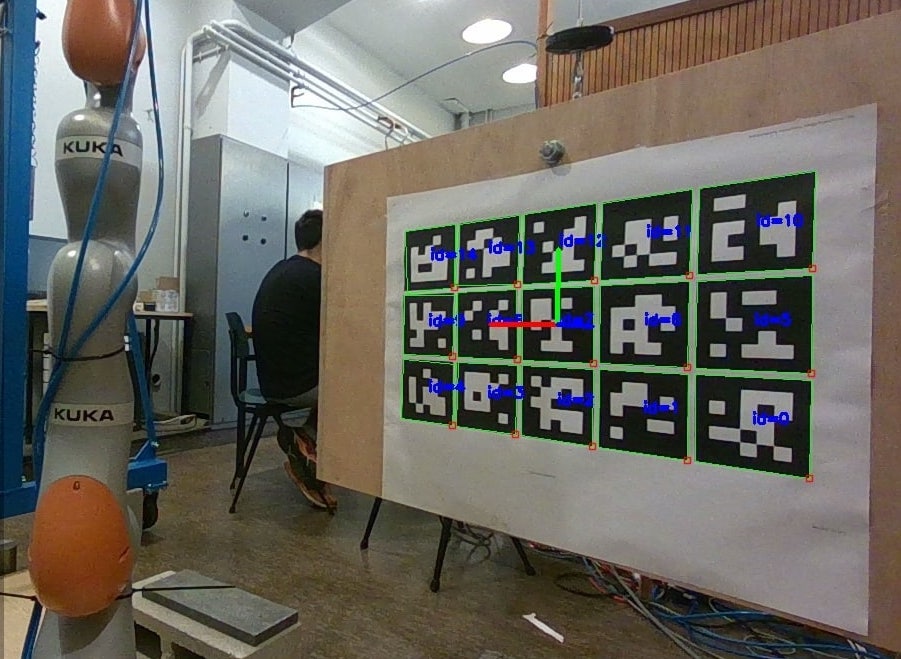}
\caption{Pose estimation of a 5x3 ArUco board frame w.r.t. camera frame  with red, green, and blue being for X, Y, Z, respectively.}
\label{ArUco_board} 
\end{figure}

For object detection and pose estimation, in this work we use identifiable fiducial markers (i.e. ArUco markers, \cite{Aruco_writers}). 
To improve the rotation estimation, an ArUco board is used instead of a single marker. A board has more markers and is more resistant to occlusions and less prone to ambiguity problems. The camera used in this work is an Intel RealSense D455 IR stereo camera. This camera is factory-calibrated, therefore the intrinsic parameters are accurately obtained from the camera's SDK and there is no need to re-preform the calibration. The computer vision library OpenCV \cite{opencv_eye2hand} is used to configure the markers and to estimate their poses, which essentially depends on the camera's intrinsic parameters and the marker's dimensions. Figure ~\ref{ArUco_board} shows the estimated pose of the board frame w.r.t. camera frame. 

\begin{figure}[H]
\centering
\includegraphics[width=0.6\columnwidth]{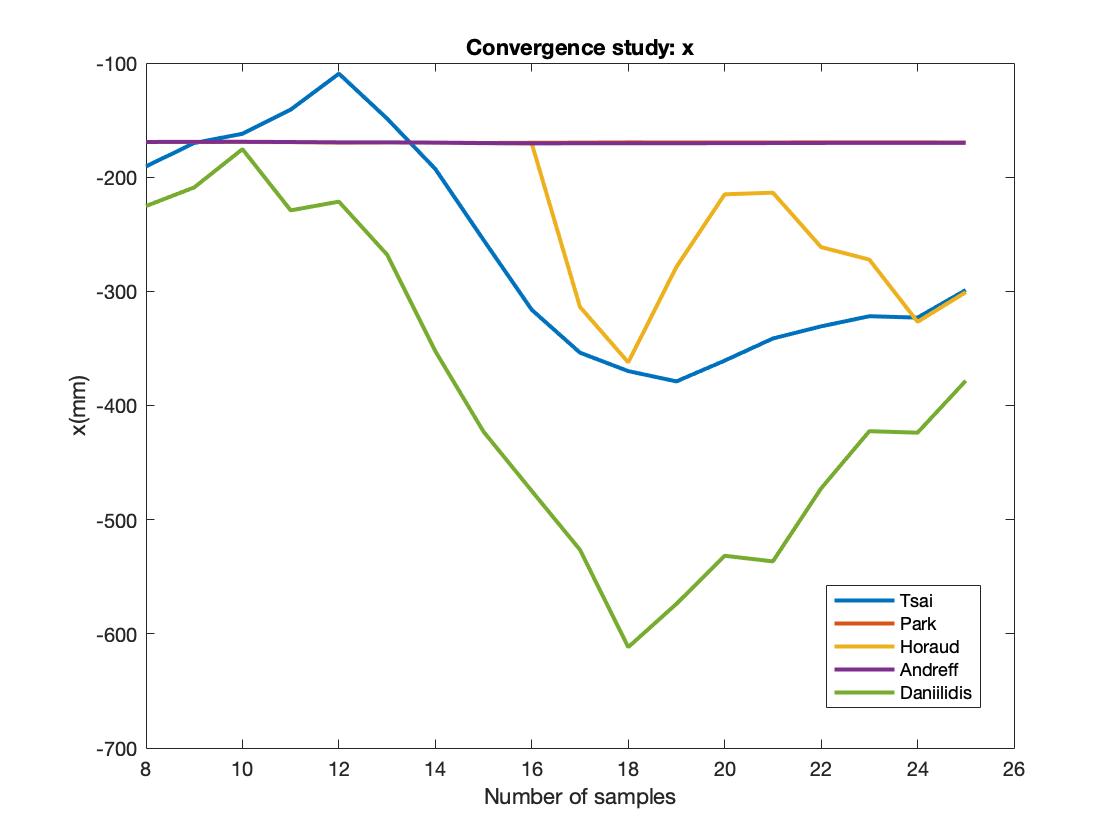}
\caption{Eye-to-hand calibration convergence study on the x-component of the obtained $T^{base}_{camera}$ \cite{Ma2014}.}
\label{conv_stud} 
\end{figure}

\begin{figure}[H]
\centering
\includegraphics[width=0.6\columnwidth]{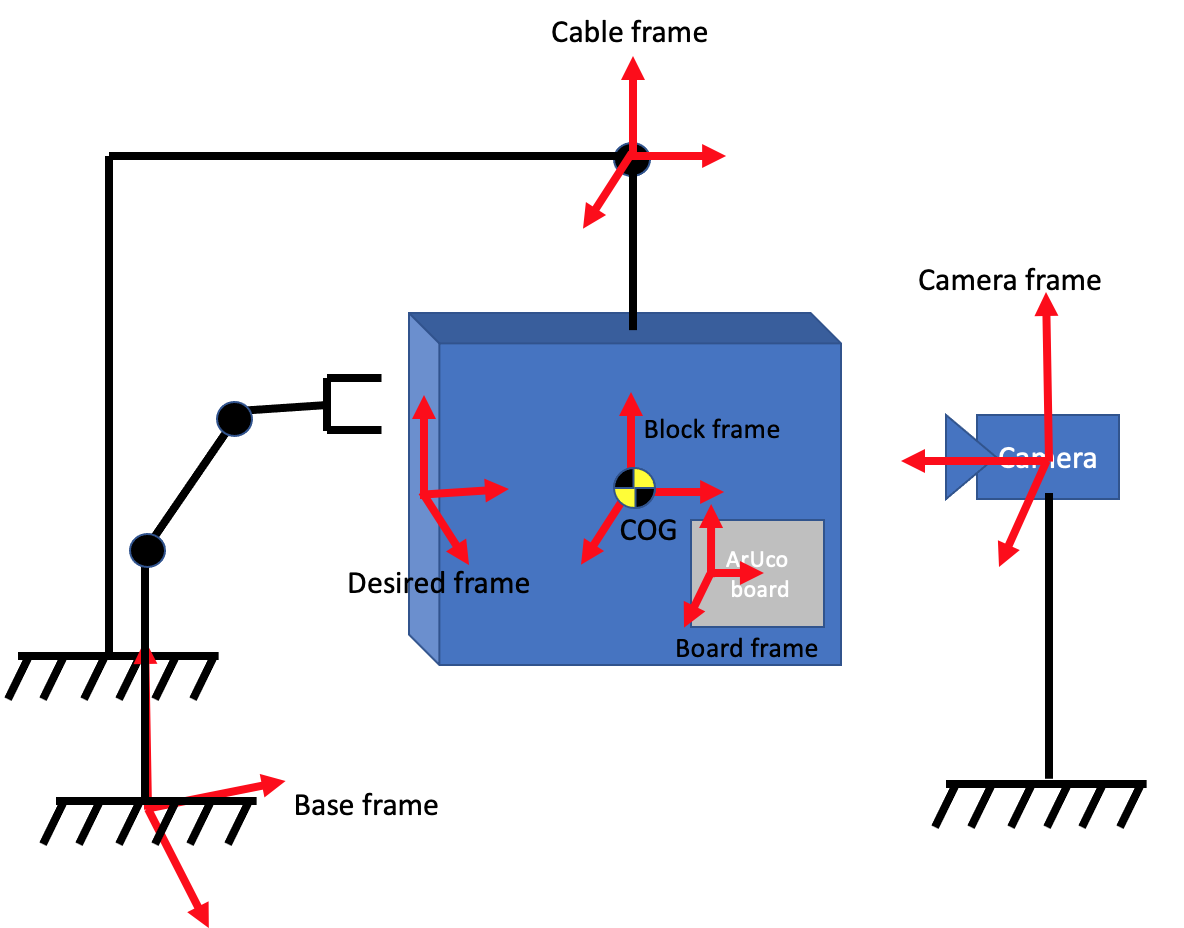}
\caption{Extrinsic calibration: scheme of all the existing reference frames. Dimensions not to scale.}
\label{frames_map} 
\end{figure}

Once the camera is fixed in a reference position, the first step is to perform the extrinsic calibration of the camera in 'eye-to-hand' configuration to align the reference frame of the camera measurements with that of the robot base frame (see Fig.~\ref{frames_map}). This choice is motivated by the fact that the controller is designed in the robot task space, therefore the desired pose to follow must be expressed in the base frame of the robot. \footnote{The purpose of this paper is not to describe this type of procedure in detail, however, a detailed guide for all the tools developed to perform and validate the calibration in this paper is present on the GitHub repository \url{https://github.com/ManarMahmalji/Eye-to-hand-Calibration-with-OpenCV.git}.} To perform the calibration, a calibration board (i.e. a ChArUco board) is rigidly attached to the gripper of the robot, and the camera is fixed. By moving the robotic arm to random positions, the problem of performing the extrinsic calibration can be algebraically seen as the solution of the system $AX=XB$, where $X$ is the homogeneous transformation that goes from the camera frame to the robot base frame (i.e. $T^{base}_{camera}$, which we are interested in), and $A$ and  $B$ can be obtained by getting, for each motion of the gripper, $T^{camera}_{board}$ (obtained directly from the camera) and $T^{robot gipper}_{base}$ (obtained from the direct kinematics of the robot), respectively. The mathematical development of this problem is well-illustrated on the website of OpenCV \cite{opencv_eye2hand}. The previously mentioned system can be solved by using different algorithms with OpenCV and a convergence analysis can be performed (see Fig.~\ref{conv_stud}). 

After this procedure, we are able to translate all the measures of the camera to the robot base frame. However, two more intermediate steps are necessary to obtain all the desired measures in the robot base frame.
\medskip

First, the camera is expected to compute the pose of a frame centered at the COG of the block (i.e. the block frame in Fig.~\ref{frames_map}) in order to evaluate the block model \eqref{eq:obj_model}. The following transformations are therefore used: 
\begin{itemize}
    \item $T^{board}_{block}$: go from  the block frame to the board frame. This can be estimated since the dimensions of the block and of the board are well-known, and it consists only of the translation vector to go from the block's COG to the board frame origin.
    \item $T^{camera}_{board}$: go from the board frame to the camera frame. This is obtained directly from the camera. 
    \item $T^{base}_{camera}$: go from the camera frame to the robot base frame. This is obtained as result of the extrinsic calibration. 
\end{itemize}

As final result we obtain:
\begin{equation}
     \small
     T^{base}_{block} =  T^{base}_{camera}  T^{camera}_{board}  T^{board}_{block}. 
\end{equation}

Second, in order to track the oscillating suspended block, the controller of the robot should receive the pose of the frame centered at the desired tracking point and oriented in the way we want the end-effector to follow the block. In this work, we use the frame on the side of the suspended object as the desired frame (i.e. the desired frame in Fig.~\ref{frames_map}). Therefore, the following transformation $T^{block}_{des}$ that goes from the desired frame to the block frame is used. This can be estimated since the dimensions of the block are known. 

\medskip

Mathematically, what is sent to the robot controller is expressed as:
\begin{equation}
     \small
     T^{base}_{des} =  T^{base}_{block} T^{block}_{des}. 
\end{equation}

A scheme of the all the referred frames is shown in Fig.~\ref{frames_map}

\section{EXTENDED KALMAN FILTER WITH INTERMITTENT OBSERVATIONS}\label{sec:EKF}

In Section~\ref{sec:probl}, we derived the dynamic model of the suspended object and assumed that the state of the system can be measured by a camera device. In practice, when using a camera device, it is necessary to handle measurement noise, possible lost packets during communication and a low data transmission rate. To solve these problems, in this section, we propose the implementation of an Extended Kalman Filter (EKF) with intermittent observations. The proposed estimator uses the dynamic model of the object and the visual information from the camera to improve the quality of the information to be sent to the robot, in order to ensure proper tracking of the object by the robot. Hereby, we briefly outline the equations that allow us to design and implement an EKF with intermittent observations. Please refer to \cite{garone2011lqg,sinopoli2004kalman} for more details on the subject.

\medskip

Discretizing \eqref{eq:Tc} with a constant sampling time $T_s$, one can obtain the following discrete state space set of equations 

\begin{equation}\label{eq:Td}
    \small
    \begin{cases}
        x_{k+1} &= f^d(x_k)+w_k \\
        y_k &= h^d(x_k)+v_k,
    \end{cases}
\end{equation}
where $w_k \in\mathbb{R}^{10}$ and $v_k\in\mathbb{R}^{7}$ are white Gaussian noise processes, uncorrelated with zero mean and covariances $Q \in\mathbb{R}^{10\times10}$ and $R \in\mathbb{R}^{7\times 7}$. 
\medskip
 
By considering \eqref{eq:Td} and assuming that the random processes $w_k$ and $v_k$ for all time step $k$ and the initial state $x_0$ are mutually independent, the implementation of an EKF with intermittent observations can be achieved by iteratively evaluating the following two steps. 

\medskip

\textit{Prediction step.} Assume that the state estimation $\hat{x}_{k|k}$ and the error covariance $P_{k|k}$ are available at time step \textit{k}, the one-step state prediction $\hat{x}_{k+1|k}$ and the one-step error covariance prediction $P_{k+1|k}$ can be estimated as  
\begin{equation}
\small
\begin{aligned}
    \hat{x}_{k+1|k} &= f^d(\hat{x}_{k|k}), \\
    P_{k+1|k} &= A_kP_{k|k}A^T_k + Q,
\end{aligned}
\end{equation}
where $A_k = \frac{\partial{f^d}}{\partial{x}}\bigg|_{x = \hat{x}_{k|k}}$ is the Jacobian matrix of the state equation evaluated around the state prediction $\hat{x}_{k|k}$.
\medskip

\textit{Correction step.} The state estimation and the error covariance at time step \textit{k+1} can be evaluated as
\begin{equation}
\small
\begin{aligned}
    \hat{x}_{k+1|k+1} &= \hat{x}_{k+1|k} + \gamma_k K_{k+1|k+1}(y_{k+1}-h^d(\hat{x}_{k+1|k}))\\
    P_{k+1|k+1} &=  P_{k+1|k}-\gamma_k K_{k+1|k+1}C_kP_{k+1|k},
\end{aligned}
\end{equation}

where $\gamma_k$ is a binary variable that will be equal to one if the measurements from the camera are available at the time instant $k$, otherwise is equal to zero, and 

\begin{equation}
\small
    K_{k+1|k+1} = P_{k+1|k}C^T_k(C_kP_{k+1|k}C^T_k + R)^{-1} 
\end{equation}
is the corresponding equation of the EKF gain, and $C_k = \frac{\partial{h^d}}{\partial{x}}\bigg|_{x = \hat{x}_{k+1|k}}$, is the Jacobian matrix of the output equation evaluated around the state prediction $\hat{x}_{k+1|k}$.
 
\subsection{EKF experimental results}

\begin{figure}[h]
\centering
\includegraphics[width=1\columnwidth]{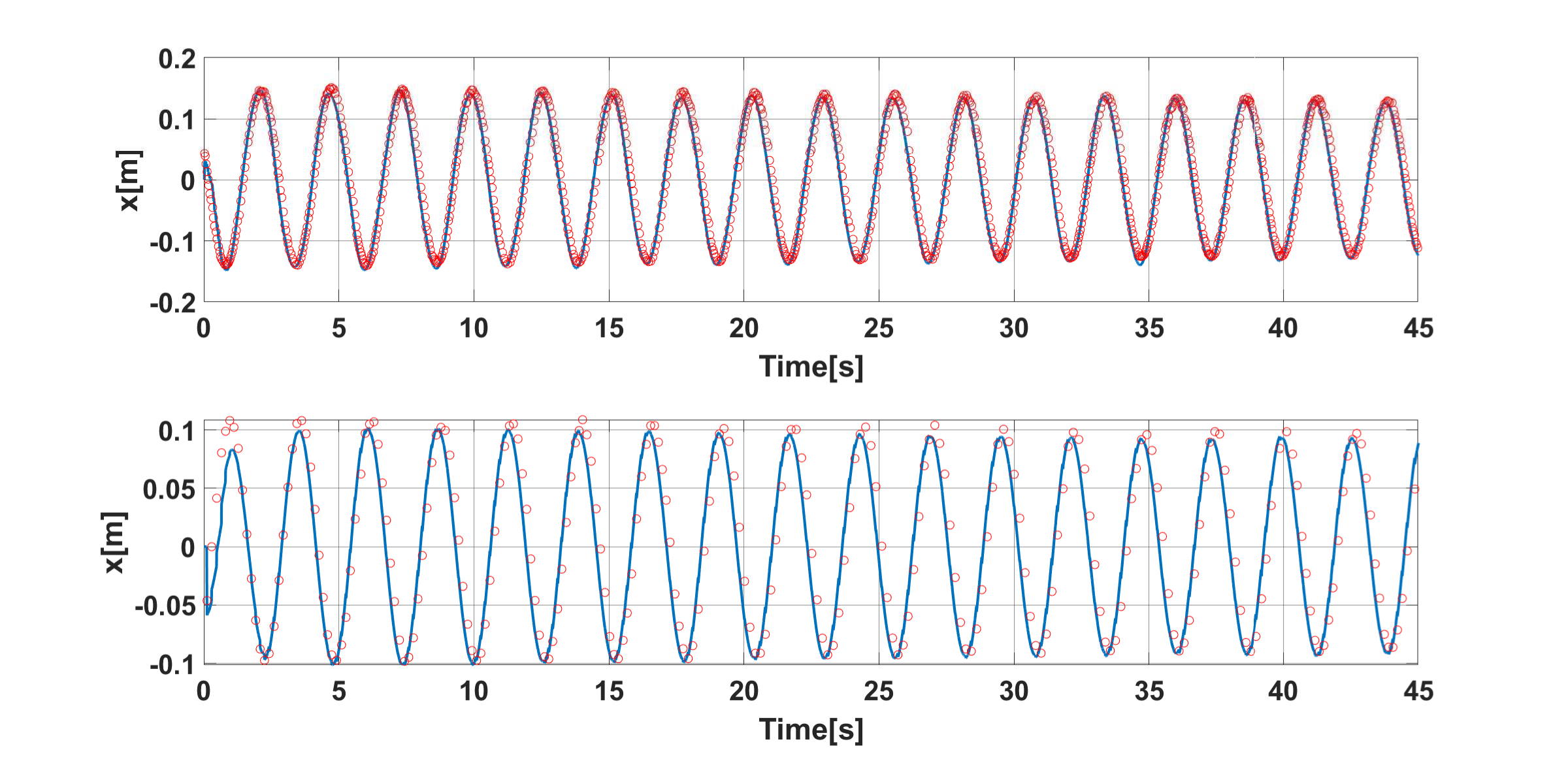}
\caption{Solid blue line: output of the EKF. Red circles: data from the camera. Upper plot: camera data rate 40Hz. Lower plot: camera data rate 10Hz.}
\label{fig:EKF1} 
\end{figure}

To demonstrate the effectiveness of the proposed EKF, two different experiments were performed, see Fig.~\ref{fig:EKF1}. In these experiments, the goal is to track the oscillation of the suspended block. The used block has a mass of 22kg and it is suspended by a cable of 1.215m. The dimensions of the block are 0.8x0.6x0.2m. The results shown in Fig.~\ref{fig:EKF1} depict an oscillation of the block that occurs mostly with respect to a displacement on the x-axis and the measurements are referred to the COG of the block itself. In the upper figure, all camera samples are used to estimate the dynamics of the block. The EKF output (blue line) follows the trend of the camera output, thus obtaining a correct estimate of the block oscillation. In the lower figure, we took one out of four samples from the camera to simulate a scenario in which packet loss can occur. As it can be seen, the designed estimator is again able to correctly estimate the dynamics of the block.

\section{EXPERIMENTS WITH THE ROBOTIC ARM}\label{sec:results}

\begin{figure}[h]
\centering
\includegraphics[width=0.8\columnwidth]{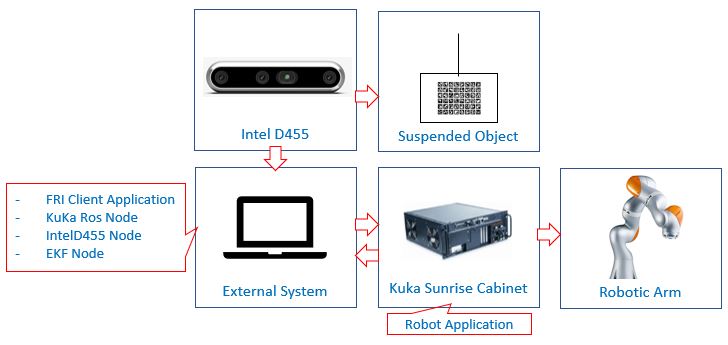}
\caption{Pipeline Architecture.}
\label{fig:pipeline} 
\end{figure}

The experiments were performed with a seven degree of freedom serial manipulator, a KUKA LBR IIWA14 R820. This robot is especially suited for research in robotics, as it is accessible through a real-time interface named \textit{Fast  Robot  Interface} (FRI). FRI is an interface via which data can be exchanged continuously and in real time between a robot application running on the robot controller and an FRI client application running on an external system. The entire hardware/software architecture is shown in Fig.~\ref{fig:pipeline}. An open source GitHub repository with the code related to the proposed architecture will be made available at \url{https://github.com/MikAmb95/EKF.git}. 

The proposed architecture comprises several elements:
\begin{itemize}
\item \textit{Kuka LBR IIWA14 R820}. It is the robotic arm used during our experiments.
\item \textit{Robot Application}. The robot application is executed on the robot controller. In this
paper, KUKA Sunrise.OS 1.17, KUKA Sunrise.Workbench 1.17 and KUKA Sunrise.FRI 1.17 are used to program the robot. 
\item \textit{Intel D455}. It is the camera device used in our experiments. The camera estimates the dynamics of the oscillating block and sends this information to an external system. 
\item \textit{External System}. In our case, it is a Laptop with Intel(R) Core(TM) i7-6500U CPU 2.50GHz 2.60 GHz. The custom nodes are executed on this system.
\item \textit{FRI Client Application}. It is created in C++. It allows sending commands to the Robot Application and receiving information on the state of the robot (e.g. read joint position sensors).
\item \textit{KuKa ROS Node}. This node implements the controller on the robot. It sends the torque commands to the KuKa Sunrise Cabinet. This node runs at  1kHz.
\item \textit{Intel D455 Node}. This node configures the camera, retrieves
raw images, estimates the pose of the block and send this information to the EKF Node. This node runs at 40 Hz.
\item \textit{EKF Node}. This node runs at 300Hz and implemented the EKF explained in Section~\ref{sec:EKF}.  
\end{itemize}

\subsection{Experimental Results}

\begin{figure}[h]
\centering
\includegraphics[width=0.5\columnwidth]{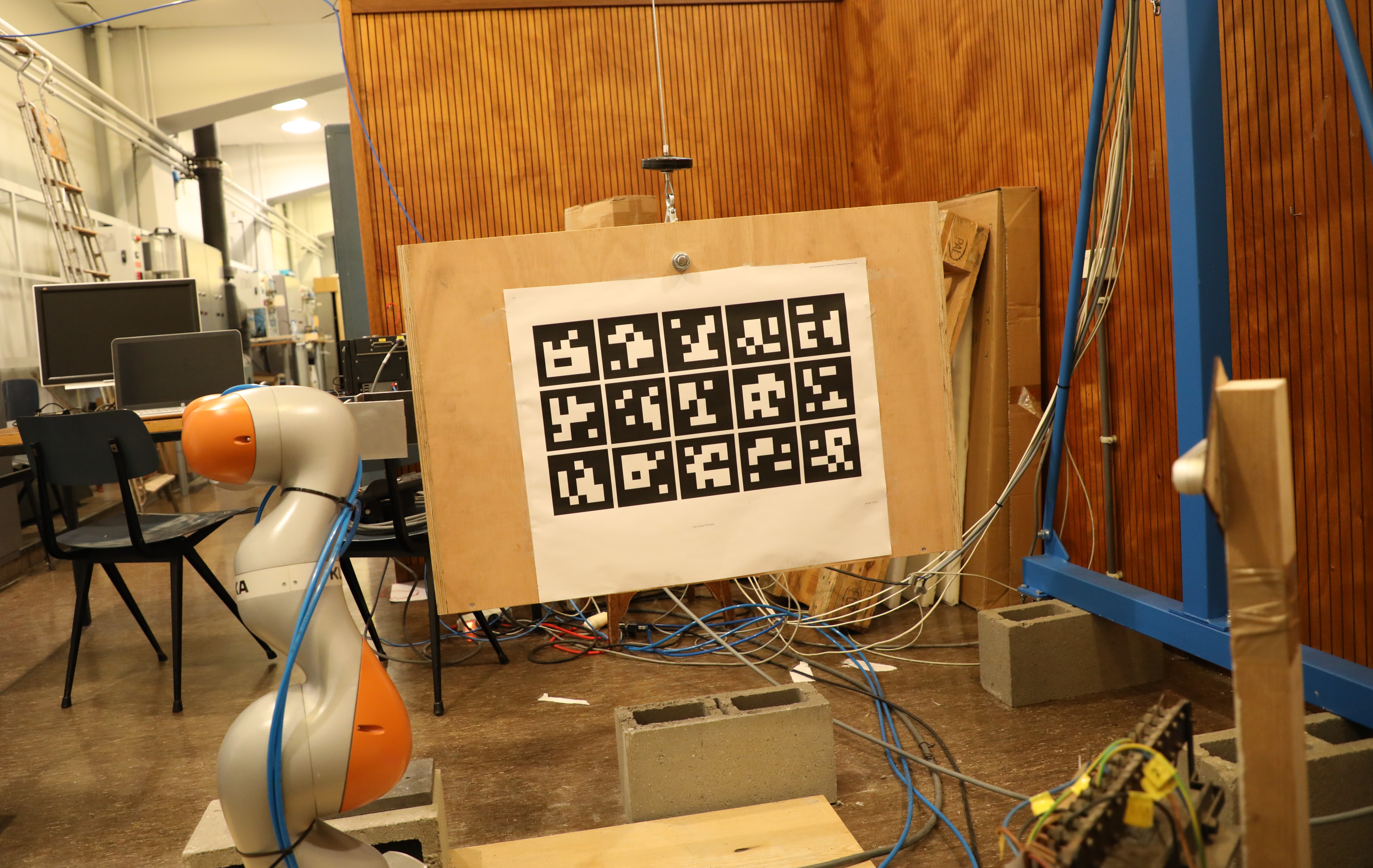}
\caption{Set-Up}
\label{fig:setup2} 
\end{figure}

\begin{figure}[h]
\centering
\includegraphics[width=0.75\columnwidth]{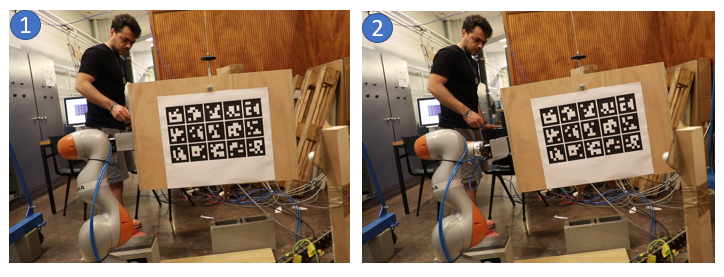}
\caption{Snapshots of the experiments.}
\label{fig:experiments} 
\end{figure}

The set-up used during the experiments is shown in Fig.~\ref{fig:setup2}. A video of the experiments can be found at \url{https://youtu.be/JXg-Y9TbjVc}. Snapshots of the experiments are reported in Fig.~\ref{fig:experiments}.

\medskip

\begin{figure}[h]
\centering
\includegraphics[width=1\linewidth]{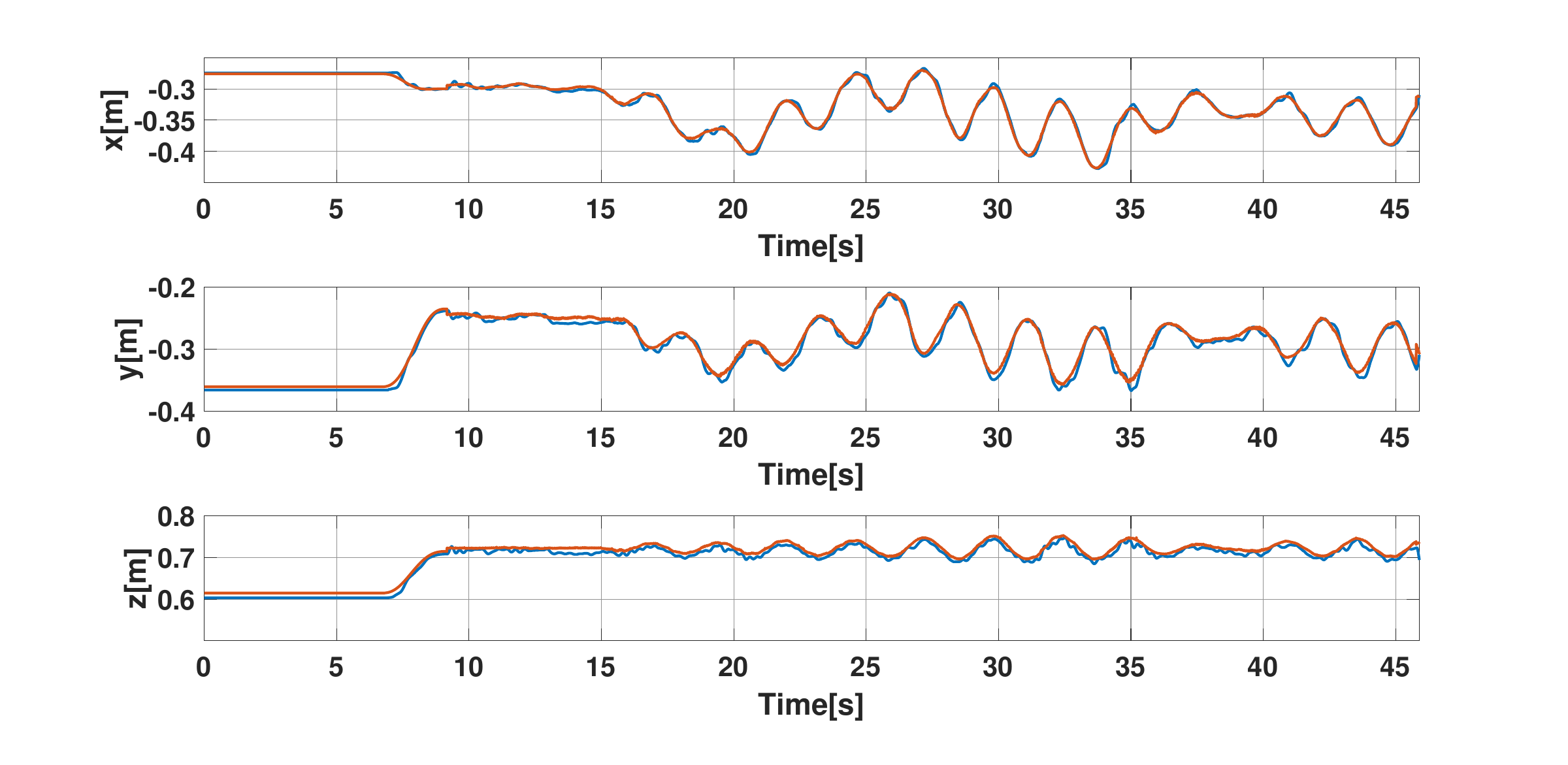}
\caption{Blue line: Current Robot end-effector position. Red line: Desired Robot end-effector position.}
\label{fig:results} 
\end{figure}

The goal of the experiments is to control the end-effector of the robot to follow the oscillations of the suspended block. In particular, the implemented estimator explained in Section~\ref{sec:EKF}, uses the dynamic model of the block and the camera data to estimate the current pose of the block (i.e. the orientation and the position of the block). Then, this pose is translated to the robot base and represents the reference for the control scheme \eqref{tau} (i.e. $x_d$), which moves the robot in such a way as to ensure correct tracking of the oscillations. The results reported in this section refer to the end-effector position of the robot (i.e. only the translation part of the vector $x_d$). As one can see in Fig.~\ref{fig:results}, the implemented EKF is able to estimate the oscillations of the block based on its dynamic model and the camera data. In addition, it reduces the noise of the measurements obtained in order to feed the robot controller with smooth signals. Thus, the controller of the robot is able to move the end-effector to properly follow the oscillations of the block.

\subsection{Implementation problems and proposed solutions}

In this work, we have addressed some implementation challenges with the aim to advance the state of the art regarding the use of visual servoing applications. 
\begin{enumerate}
    \item KDL Library \cite{KDL}. Most of the results present in the scientific community make use of the Kinematics and Dynamics Library (KDL), which develops a framework for modeling and calculating kinematic chains, such as robots. However, for real-time applications, the use of this library can affect performance by being very computationally heavy. The work presented in this paper does not make use of this library, but a new approach was implemented using an alternative symbolic tool. More information can be found in the two GitHub repositories linked to this paper.   
    \item Data Rate. The main problem in a vision servoing application is the low data rate coming from the camera device. The main contribution of this work is the proposed EKF which helps to increase the data rate and counteract the negative effects of measurement noise 
    \item Use of the own controller of the robot. Typically, the algorithm that controls the robot is implemented in its internal controller (in the Robot Application), seen in Fig.~\ref{fig:pipeline}. However, this approach can introduce unwanted delays during the following of a desired reference. As one can see in Fig.~\ref{fig:results2}, in case \eqref{tau} is implemented in the Robotic Architecture level, there is a delay between the desired reference that comes from the EKF and the motion of the robot. In this paper we solved this problem by implementing \eqref{tau} in the external system as C++ application.
    \begin{figure}[h]
\centering
\includegraphics[width=1\linewidth]{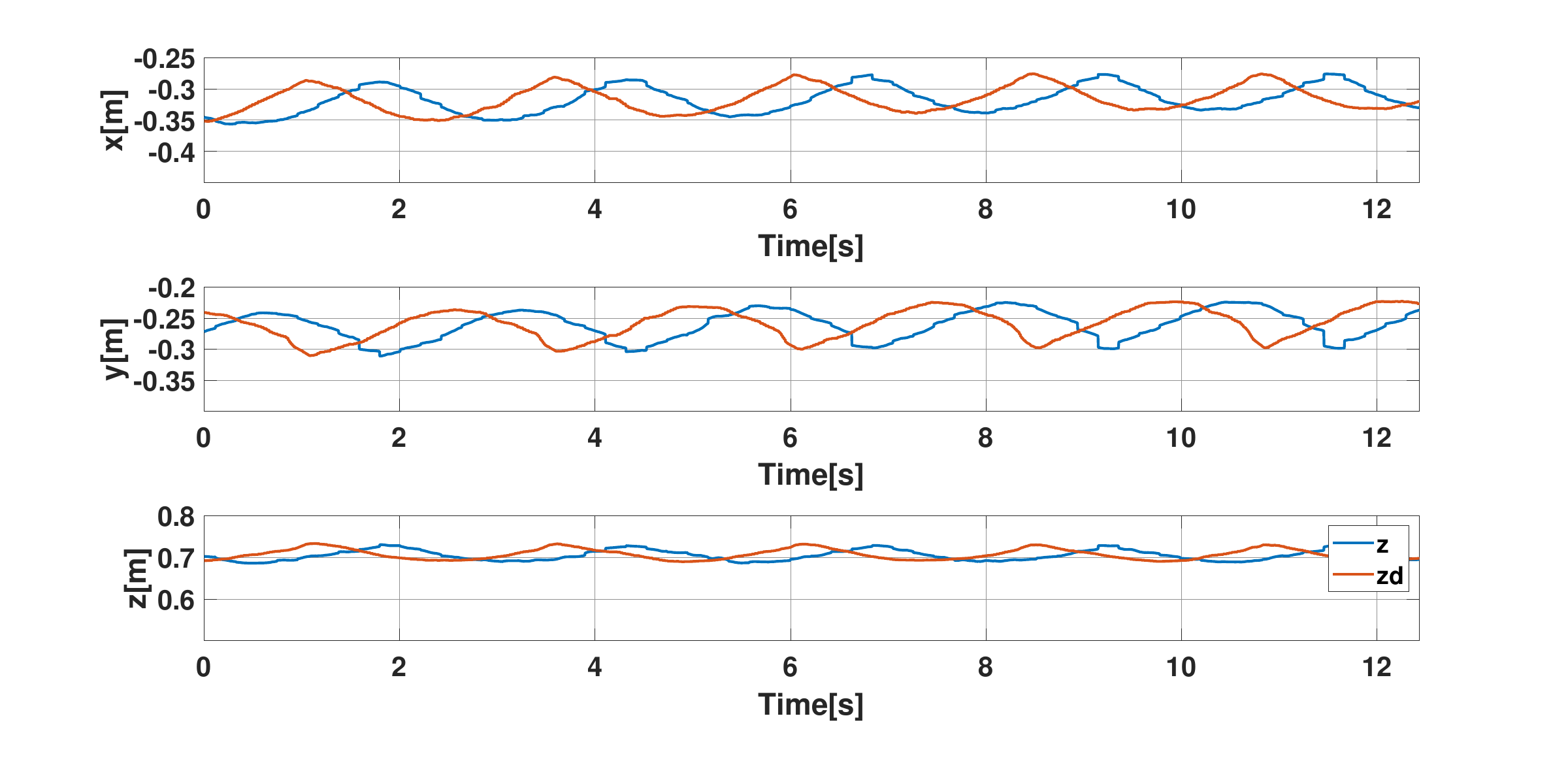}
\caption{Blue line: Current Robot end-effector position. Red line: Desired Robot end-effector position.}
\label{fig:results2} 
\end{figure}
    
\end{enumerate}

\section{CONCLUSIONS}\label{sec:conc}
In this paper, we implemented an Extended Kalman Filter with Intermittent Observation to track the oscillations of a suspended object with a robotic arm. In particular, the proposed estimator uses the dynamic model of the target object and the visual information from a camera device in an eye-to-hand configuration, to feed the controller of the robot with a smooth desired reference. By using the EKF, we were able to solve common problems of visual servoing applications (e.g. limited data transmission rates, noisy measurements) that in the case of a moving object could affect the control of the robot itself. Future works will use the present set-up and the implemented EKF to perform also the grabbing of the suspended object.

\bibliographystyle{IEEEtran}
\bibliography{main}

\end{document}